\documentclass{article} 
\usepackage{iclr2015,times}
\usepackage{hyperref}
\usepackage{url}

\usepackage{bm}
\usepackage{color}
\usepackage{graphicx}
\usepackage{amsmath}
\usepackage{amssymb}
\usepackage{subfigure} 
\usepackage{adjustbox}

\title{Zero-bias autoencoders and the benefits of co-adapting features}

\author{
Kishore Konda\\
Goethe University Frankfurt\\
Germany\\
\texttt{konda.kishorereddy@gmail.com}\\
\And
Roland Memisevic\\
University of Montreal\\
Canada\\
\texttt{roland.memisevic@umontreal.ca} \\
\And
David Krueger\\
University of Montreal\\
Canada\\
\texttt{david.krueger@umontreal.ca}\\
}

\iclrfinalcopy 

\iclrconference 

\begin{document}

\maketitle

\begin{abstract}
Regularized training of an autoencoder typically results in hidden unit biases that take on large negative values. We show that negative biases are a natural result of using a hidden layer whose responsibility is to both represent the input data and act as a selection mechanism that ensures sparsity of the representation. We then show that negative biases impede the learning of data distributions whose intrinsic dimensionality is high. We also propose a new activation function that decouples the two roles of the hidden layer and that allows us to learn representations on data with very high intrinsic dimensionality, where standard autoencoders typically fail. Since the decoupled activation function acts like an implicit regularizer, the model can be trained by minimizing the reconstruction error of training data, without requiring any additional regularization. 
\end{abstract}

\section{Introduction}
Autoencoders are popular models used for learning features and pretraining deep networks.  
In their simplest form, they are based on minimizing the squared error between 
an observation, $\bm{x}$, and a non-linear reconstruction defined as 
\begin{equation}
\bm{r}({\bm{x}}) = \sum_k h\big( \bm{w}_{k}^\mathrm{T}\bm{x} + b_k \big) \bm{w}_k + \bm{c}
\label{eq:AE}
\end{equation}
where $\bm{w}_k$ and $b_k$ are weight vector and bias for hidden unit $k$, 
$\bm{c}$ is a vector of visible biases, and $h(\cdot)$ is a hidden unit activation 
function. 
Popular choices of activation function are the sigmoid $h(a)=\big(1+\exp(-a)\big)^{-1}$, 
or the rectified linear (ReLU) $h(a)=\max(0,a)$. 
Various regularization schemes can be used to prevent trivial solutions when 
using a large number of hidden units. These include 
corrupting inputs during learning \cite{denoisingAE}, 
adding a ``contraction'' penalty which forces derivatives of hidden unit activations 
to be small \cite{contractiveAE}, or using sparsity penalties \cite{coatessinglelayer}.

This work is motivated by the empirical observation that across a wide range of applications, 
hidden biases, $b_k$, tend to take on large negative values when training an autoencoder 
with one of the mentioned regularization schemes. 

In this work, we show that negative hidden unit biases are at odds with some 
desirable properties of the representations learned by the autoencoder.  
We also show that negative biases are a simple consequence of the fact that hidden units  
in the autoencoder have the dual function of (1) selecting which weight vectors take part in 
reconstructing a given training point, 
and (2) representing the coefficients with which the selected weight vectors get combined 
to reconstruct the input (cf., Eq.~\ref{eq:AE}).

To overcome the detrimental effects of negative biases, 
we then propose a new activation function that allows us to disentangle these roles. 
We show that this yields features that increasingly outperform regularized autoencoders 
in recognition tasks of increasingly high dimensionality. 
Since the regularization is ``built'' into the activation function, it allows us 
to train the autoencoder without additional regularization, like contraction or 
denoising, by simply minimizing reconstruction error. 
We also show that using an encoding without negative biases at test-time in both 
this model and a contractive autoencoder achieves state-of-the-art performance on 
the permutation-invariant CIFAR-10 
dataset.\footnote{An example implementation of the zero-bias autoencoder in python is 
available at \url{http://www.iro.umontreal.ca/~memisevr/code/zae/}.}

\subsection{Related work}
Our analysis may help explain why in a network with linear hidden units, the optimal 
number of units tends to be relatively small \cite{ba2013adaptive,ksparse}. 
Training via thresholding, which we introduce in Section~\ref{section:thresholding}, 
is loosely related to dropout \cite{dropout}, in that it forces features to align with 
high-density regions. In contrast to dropout, our thresholding scheme is not stochastic. 
Hidden activations and reconstructions are a deterministic function of the input. 
Other related work is the work by \cite{goroshin-lecun-iclr-13} who introduce a variety of new 
activation functions for training autoencoders and argue for 
shrinking non-linearities (see also \cite{hyvarinen2004independent}), 
which set small activations to zero. 
In contrast to that work, we show that it is possible 
to train autoencoders without additional regularization,  
when using the right type of shrinkage function. 
Our work is also loosely related to \cite{MartensRBMefficiency} who discuss limitations of 
RBMs with \emph{binary} observations.

\section{Negative bias autoencoders}
\label{section:negbias}
\begin{figure}[t]
\begin{tabular}{cc}
\begin{adjustbox}{valign=t}
\begin{tabular}{cc}
  \includegraphics[width=0.20\linewidth]{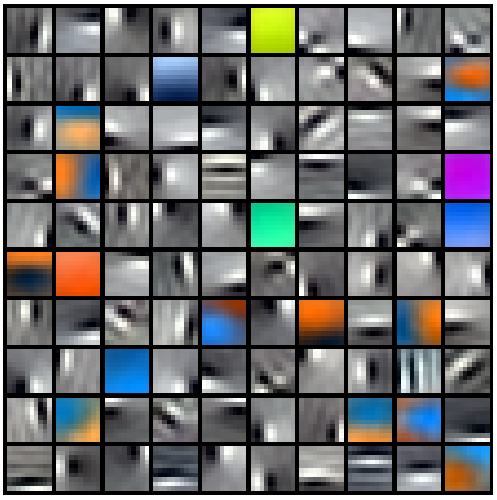}
& \includegraphics[width=0.20\linewidth]{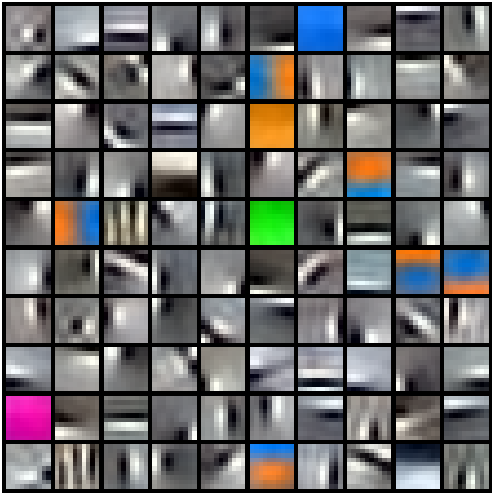} \\
  \includegraphics[width=0.20\linewidth]{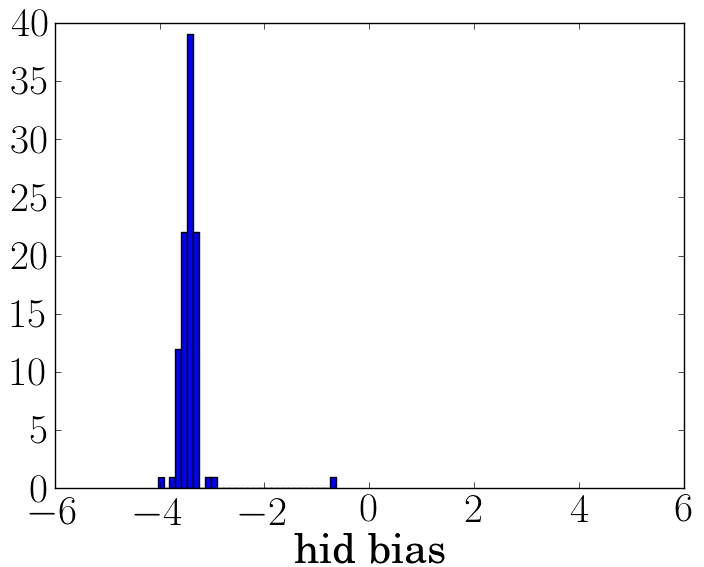}
& \includegraphics[width=0.22\linewidth]{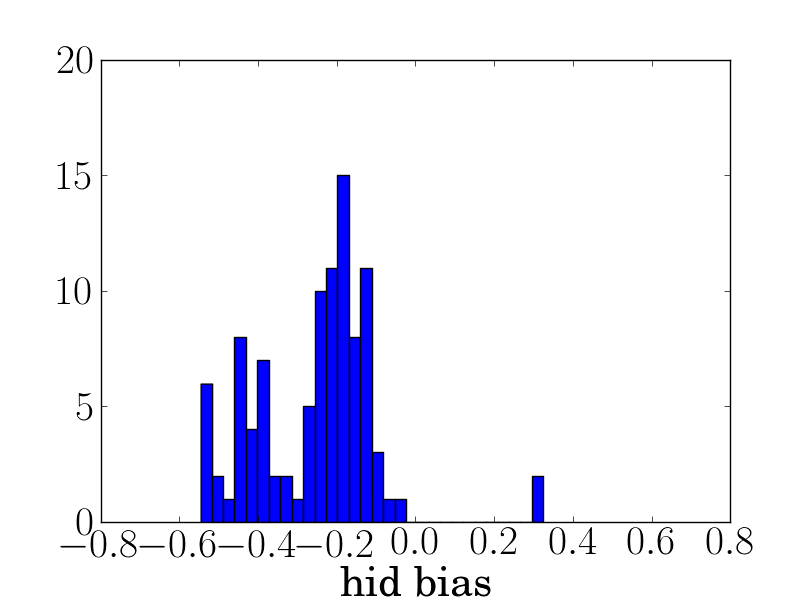}
\end{tabular}
\end{adjustbox}
&
\begin{adjustbox}{valign=t}
\includegraphics[width=0.5\columnwidth]{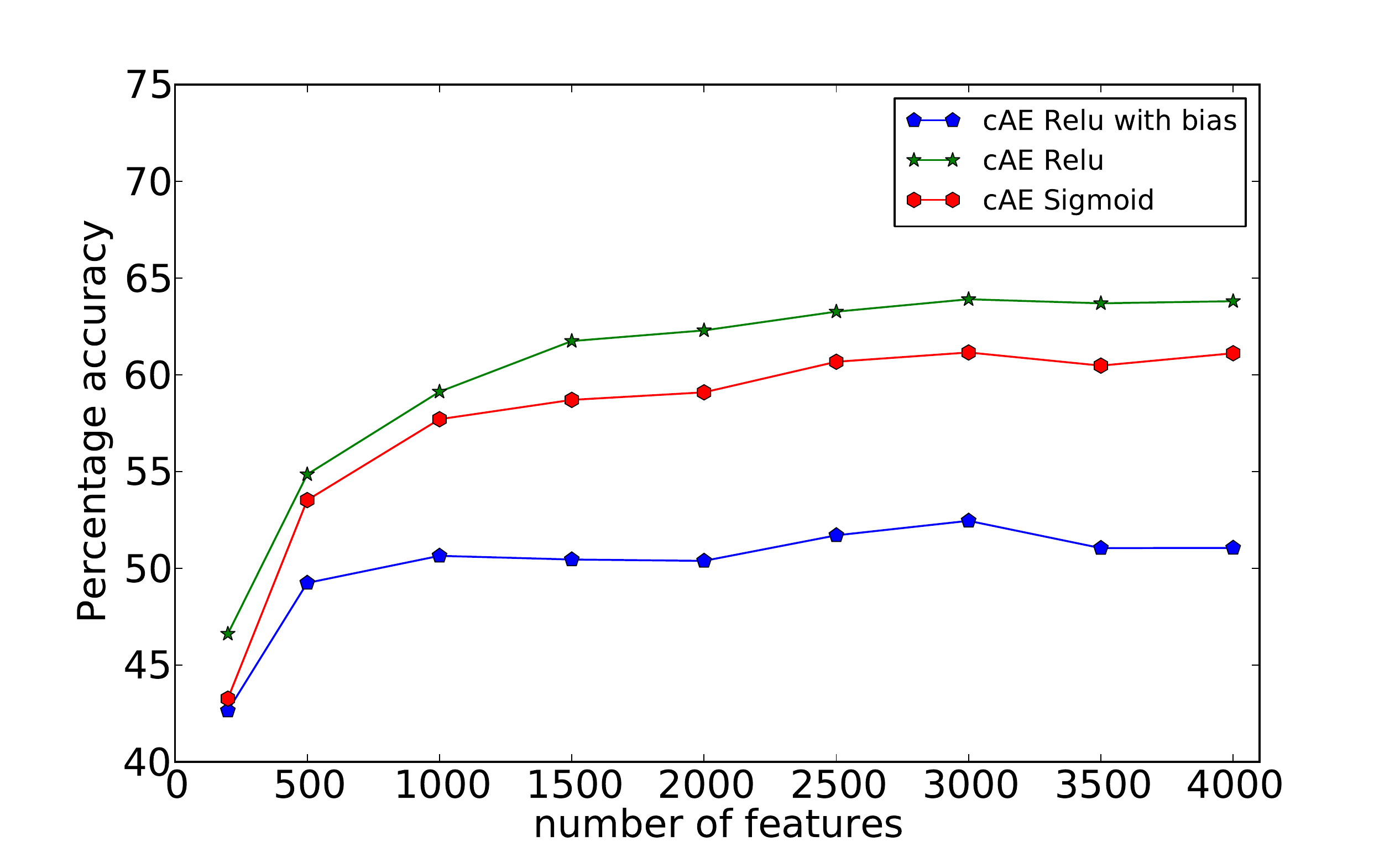}
\end{adjustbox}
\end{tabular}
\caption{Left: Filters learned by a sigmoid contractive autoencoder \cite{contractiveAE}
(contraction strength $1.0$; left) and 
a ReLU denoising autoencoder \cite{denoisingAE} (zeromask-noise 0.5; right) 
from CIFAR-10 patches, and resulting histograms 
over learned hidden unit biases. 
Right: Classiﬁcation accuracy on permutation invariant CIFAR-10 data using cAE 
with multiple different inference schemes.
All plots in this paper are best viewed in color.
}
\label{figure:negbiases}
\end{figure}

This work is motivated by the observation that regularized training of most common 
autoencoder models tends to yield hidden unit biases which are negative. 
Figure \ref{figure:negbiases} shows an experimental demonstration of this effect using 
whitened $6\times6$-CIFAR-10 color patches \cite{krizhevsky2009learning}.  
Negative biases and sparse hidden units have also been shown to be important for obtaining 
good features with an RBM 
\cite{LeeSparseRBM,HintonPracticalGuideRBM}. 

\subsection{Negative biases are required for learning and bad in the encoding}
Negative biases are arguably important for training autoencoders, especially 
overcomplete ones, because they help constrain capacity and localize features. 
But they can have several undesirable consequences on the encoding as we shall discuss. 

Consider the effect of a negative bias on a hidden 
unit with ``one-sided activation functions'', such as ReLU or sigmoid 
(i.e. activations which asymptote at zero for increasingly negative preactivation): 
On contrast-normalized data, 
it will act like a selection function, zeroing 
out the activities for points whose inner product with the weight vector $\bm{w}_k$ is small. 
As a result, the region on the hypersphere that activates a hidden unit (ie. that yields a value 
that is significantly different from $0$) will be a spherical cap, whose size is determined 
by the size of the weight vector and the bias. 
When activations are defined by spherical caps, the model 
effectively defines a radial basis function network on the hypersphere. 
(For data that is not normalized, it will still have the effect of limiting the number of training examples for which the activation function gets active.)



As long as the regions where weight vectors become active do not overlap this will 
be equivalent to clustering. In contrast to clustering, regions may of course overlap 
for the autoencoder. However, as we show in the following on the basis of an autoencoder 
with ReLU hidden units and negative biases, even where active regions merge, the model 
will resemble clustering, in that it will learn a point attractor to represent that region. 
In other words, the model will not be able to let multiple hidden units ``collaborate'' to 
define a multidimensional region of constant density. 



\subsubsection{Modes of the density learned by a ReLU autoencoder}
We shall focus on autoencoders with ReLU activation function in the following. We add an approximate argument about sigmoid autoencoders in Section~\ref{subsubsection:sigmoid} below. 

Consider points $\bm{x}$ with $r(\bm{x}) = \bm{x}$ which can be reconstructed perfectly 
by the autoencoder. 
The set of such points may be viewed as the mode of the true data generating density, 
or the true ``manifold'' the autoencoder has to find. 

For an input $\bm{x}$, define the \emph{active set} (see also \cite{numberofregions}) 
as the set of hidden units which yield a positive response: 
$
S(\bm{x})=\{k:\bm{w}_k^\mathrm{T}\bm{x}+b_k>0\}.
$
Let $W_{S(\bm{x})}$ denote the weight matrix restricted to the active units. 
That is, $W_{S(\bm{x})}$ contains in its columns the weight vectors 
associated with active hidden units for data point $\bm{x}$.
The fixed point condition $r(\bm{x}) = \bm{x}$ for the ReLU autoencoder can now be written
\begin{equation}
W_{S(\bm{x})}(W_{S(\bm{x})}^\mathrm{T}\bm{x} + \bm{b}) = \bm{x}, 
\label{eq:perfectreconstruction}
\end{equation}
or equivalently,  
\begin{equation}
(W_{S(\bm{x})}W_{S(\bm{x})}^\mathrm{T}-I) \bm{x} = - W_{S(\bm{x})} \bm{b} 
\end{equation}
This is a set of inhomogeneous linear equations, whose solutions are given by a 
specific solution plus the null-space of $M=(W_{S(\bm{x})}W_{S(\bm{x})}^\mathrm{T}-I)$. 
The null-space is given by the eigenvectors corresponding to the unit eigenvalues of 
$W_{S(\bm{x})}W_{S(\bm{x})}^\mathrm{T}$. The 
number of unit eigenvalues is equal to the number of orthonormal weight vectors in $W_{S(\bm{x})}$. 

Although minimizing the reconstruction error without bias, 
$
\|\bm{x} - W_{S(\bm{x})}W^\mathrm{T}_{S(\bm{x})} \bm{x} \|^2
$, 
would enforce orthogonality of $W_{S(\bm{x})}$ for those hidden units that are active together, 
learning with a fixed, non-zero $\bm{b}$ will not: 
it amounts to minimizing the reconstruction error between $\bm{x}$ and a shifted projection, 
$
\|\bm{x} - W_{S(\bm{x})} W^\mathrm{T}_{S(\bm{x})} \bm{x} + W_{S(\bm{x})} \bm{b}\|^2 
$, 
for which the orthonormal solution is no longer optimal (it has to account for the 
non-zero translation $W_{S(\bm{x})} \bm{b}$). 

\subsubsection{Sigmoid activations}
\label{subsubsection:sigmoid}
The case of a sigmoid activation function is harder to analyze because the sigmoid 
is never exactly zero, and so the notion of an active set cannot be used. 
But we can characterize the manifold learned by a sigmoid autoencoder (and thereby 
an RBM, which learns the same density model (\cite{aescoring})
approximately using the binary activation 
$h_k(\bm{x})=\big((\bm{w}^\mathrm{T}_k \bm{x}) + b_k \ge 0 \big)$. The reconstruction 
function in this case would be 
$$
r(\bm{x})=\sum_{k: \big((\bm{w}^\mathrm{T}_k \bm{x}) + b_k \ge 0 \big)} w_k
$$
which is simply the superposition of active weight vectors 
(and hence not a multidimensional manifold either). 

\subsubsection{Zero-bias activations at test-time}
This analysis suggests that even though negative biases are required to achieve sparsity, 
they may have a detrimental effect in that they make it difficult to learn a non-trivial manifold. 
The observation that sparsity can be detrimental 
is not new, and has already been discussed, for example, 
in \cite{ranzato2007unsupervised,kavukcuoglu2010fast}, where
the authors give up on sparsity at test-time and show that this improves recognition performance. 
Similarly, \cite{coatessinglelayer,ICML2011Saxe_551} showed that 
very good classification performance can be achieved using a linear classifier applied to 
a bag of features, using ReLU activation without bias. 
They also showed how this classification scheme is 
robust wrt. the choice of learning method used for obtaining features (in fact,
it even works with random training points as features, or using K-means as the feature 
learning method).\footnote{In \cite{coatessinglelayer} the so-called ``triangle activation'' 
was used instead of a ReLU as the inference method for $K$-means. 
This amounts to setting activations below the mean activation to zero, and 
it is almost identical to a zero-bias ReLU since the mean linear preactivation 
is very close to zero on average.} 

In Figure~\ref{figure:negbiases} we confirm this finding, and we show that it 
is still true when 
features represent whole CIFAR-10 images (rather than a bag of features).
The figure shows the classification performance of a standard contractive autoencoder 
with sigmoid hidden units trained on the permutation-invariant CIFAR-10 training 
dataset (ie. using the whole images not patches for training), 
using a linear classifier applied to the hidden activations. 
It shows that much better classification performance (in fact better than the previous 
state-of-the-art in the permutation invariant task) is achieved when replacing the sigmoid activations 
used during training with a zero-bias ReLU activation at test-time 
(see Section~\ref{sec:RectlinInf} for more details).

\section{Learning with threshold-gated activations}
\label{section:thresholding}
%
%

In light of the preceding analysis, hidden units should promote sparsity during learning, 
by becoming active in only a small region of the input space, but once a hidden unit is 
active it should use a \emph{linear} not affine encoding. 
Furthermore, any sparsity-promoting process should be removed at test time. 

To satisfy these criteria 
we suggest separating the \emph{selection} function, which sparsifies hiddens, from 
the \emph{encoding}, which defines the representation, and should be linear. 
To this end, we define the autoencoder reconstruction as the product of the selection 
function and a linear representation: 
\begin{equation}
\bm{r}(\bm{x}) = \sum_k h\big( \bm{w}_k^\mathrm{T}\bm{x} \big) \big( \bm{w}_k^\mathrm{T}\bm{x} \big) \bm{w}_k 
\label{eq:trec}
\end{equation}
The selection function, $h(\cdot)$, may use 
a negative bias to achieve sparsity, but once active, a hidden unit uses 
a linear activation to define the coefficients in the reconstruction.
This activation function is reminiscent of spike-and-slab models 
(for example, \cite{courville2011spike}), 
which define probability distributions over hidden variables as the product of a 
binary spike variable and a real-valued code. In our case, the product does not come with 
a probabilistic interpretation and it only serves to define a deterministic 
activation function which supports a linear encoding. 
The activation function is differentiable almost everywhere, so 
one can back-propagate through it for learning. 
The activation function is also related to adaptive dropout (\cite{ba2013adaptive}), 
which however is not differentiable and thus cannot be trained with back-prop. 

\subsection{Thresholding linear responses}
\label{subsection:thresholdgating}
\begin{figure}[t]
\begin{center}
\begin{tabular}{cc}
\resizebox{6.0cm}{!}{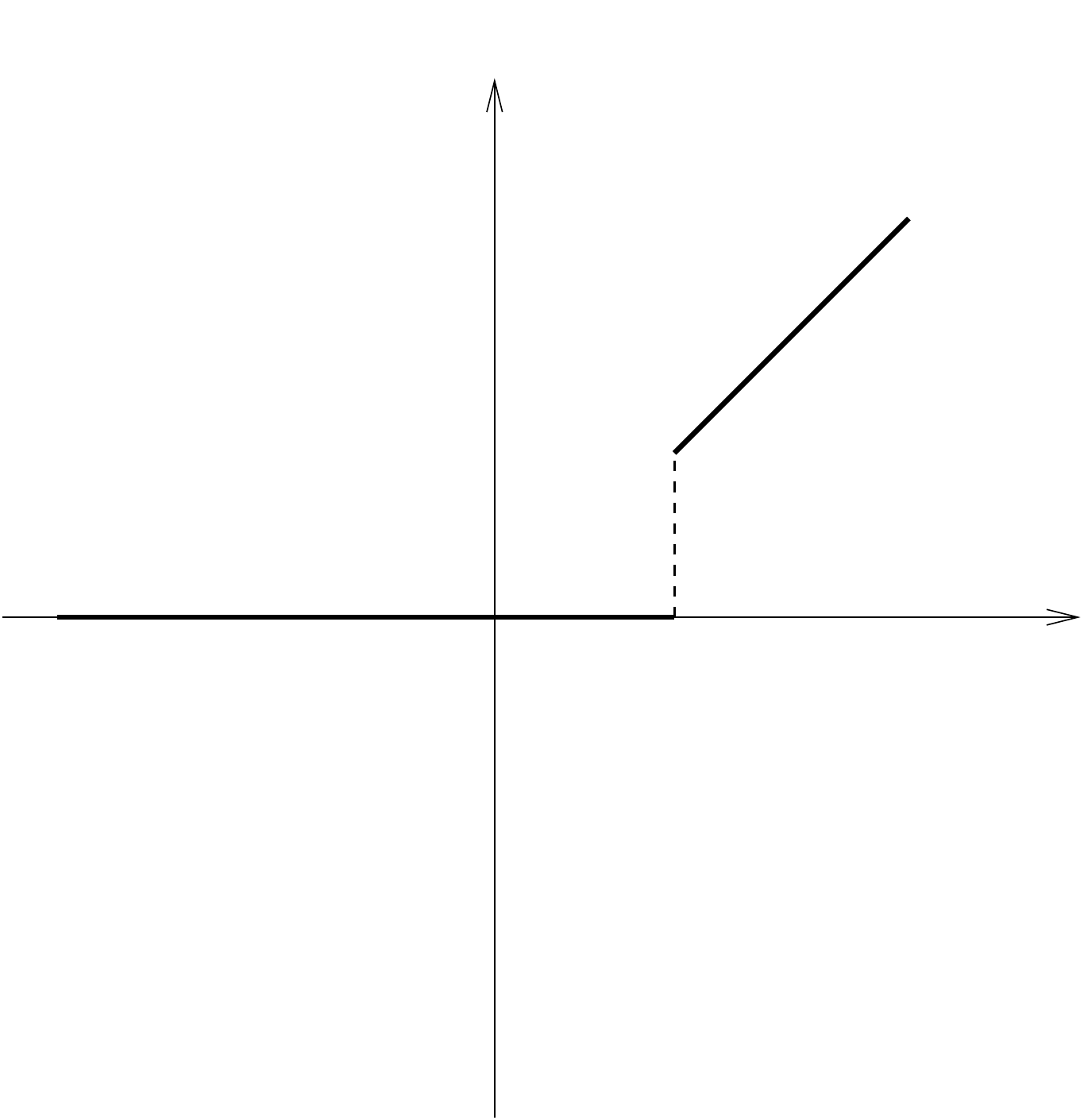} 
&
\resizebox{6.0cm}{!}{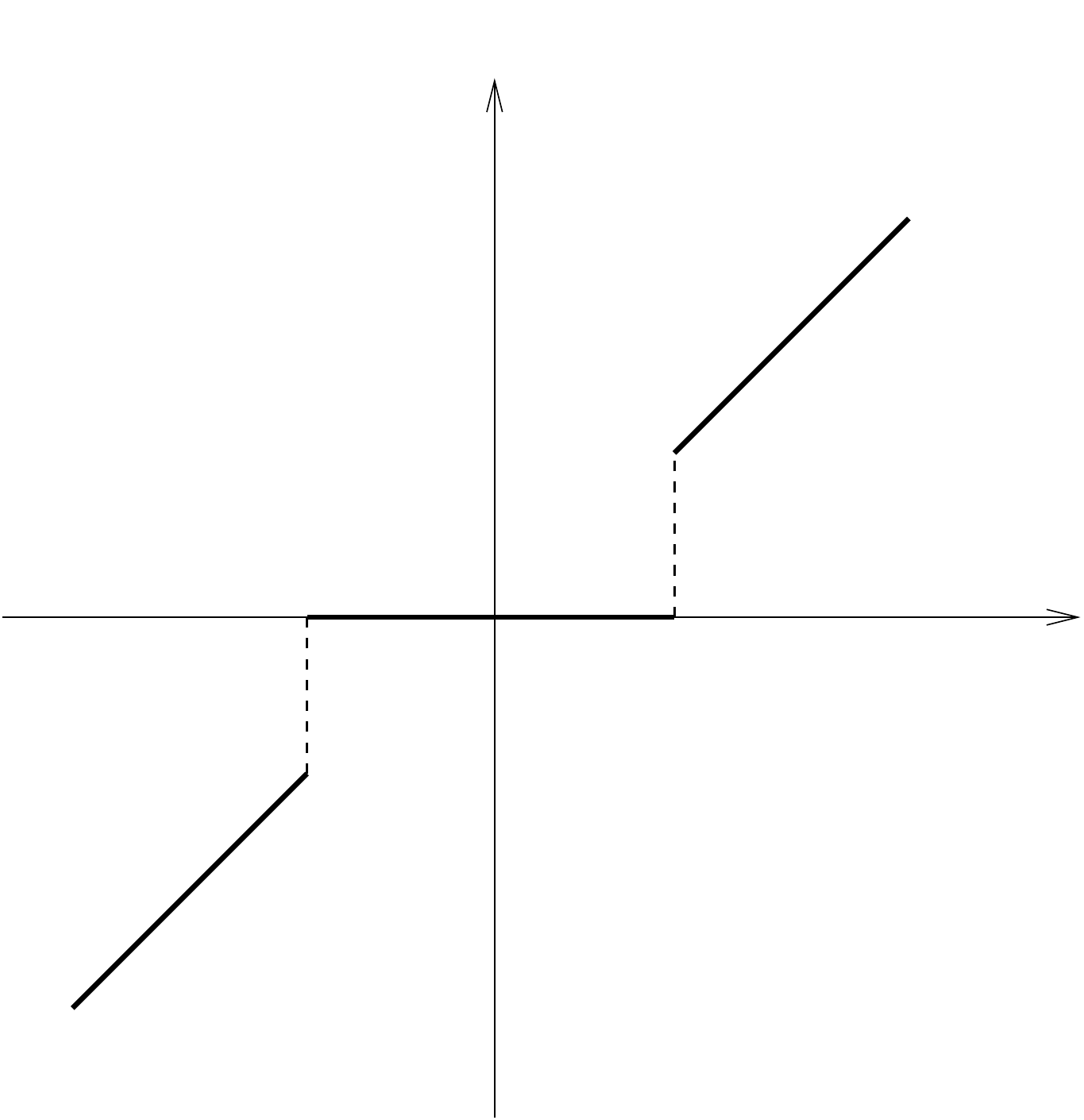} 
\end{tabular}
\caption{Activation functions for training autoencoders: 
thresholded rectified (left); thresholded linear (right).}
\label{figure:thresholdedactivations}
\end{center}
\end{figure}

In this work, we propose as a specific choice for $h(\cdot)$ the boolean selection function
\begin{equation}
h\big( \bm{w}_k^\mathrm{T}\bm{x} \big)=\big( \bm{w}_k^\mathrm{T}\bm{x} > \theta \big)
\end{equation}
With this choice, the overall activation function is 
$\big(\bm{w}_k^\mathrm{T}\bm{x}>\theta)\bm{w}_k^\mathrm{T}\bm{x}$.  
It is shown in Figure~\ref{figure:thresholdedactivations} (left). 
From the product rule, and the fact that the derivative of the boolean expression 
$\big(\bm{w}_k^\mathrm{T}\bm{x}>\theta)$ is zero, it follows that the derivative 
of the activation function wrt. to the unit's net 
input, $\bm{w}_k^\mathrm{T}\bm{x}$, is 
$\big(\bm{w}_k^\mathrm{T}\bm{x}>\theta) \cdot 1 + 0 \cdot \bm{w}_k^\mathrm{T}\bm{x}
=\big(\bm{w}_k^\mathrm{T}\bm{x}>\theta) $.
Unlike for ReLU, the non-differentiability of the activation function at $\theta$ 
is also a non-continuity. As common with ReLU activations, we train with (minibatch) 
stochastic gradient descent and ignore the non-differentiability during the optimization.  

We will refer to this activation function as Truncated Rectified (TRec) in the following. 
We set $\theta$ to $1.0$ in most of our experiments (and all hiddens have the same threshold). 
While this is unlikely to be optimal, we found it to work well and 
often on par with, or better than, traditional regularized autoencoders like the 
denoising or contractive autoencoder.
Truncation, in contrast to the negative-bias ReLU, can also be viewed as a 
hard-thresholding operator, the inversion of which is fairly 
well-understood \cite{RecoveryFromThresholded}. 

Note that the TRec activation function is simply a peculiar activation function that we 
use for training. So training amounts to minimizing squared error without any kind 
of regularization. 
We drop the thresholding for testing, where we use simply the rectified linear response. 

We also experiment with autoencoders that use a ``subspace'' variant of the TRec 
activation function (\cite{rozell2008sparse}), given by 
\begin{equation}
\bm{r}(\bm{x}) = \sum_k h\big( (\bm{w}_k^\mathrm{T}\bm{x})^2 \big) \big( \bm{w}_k^\mathrm{T}\bm{x} \big) \bm{w}_k 
\label{eq:tlin}
\end{equation}
It performs a linear reconstruction when the preactivation is either very large or 
very negative, so 
the active region is a subspace rather than a convex cone. 
To contrast it with the rectified version, we refer to this activation function as 
thresholded linear (TLin) below, but it is also known as hard-thresholding in the 
literature \cite{rozell2008sparse}. 
See Figure~\ref{figure:thresholdedactivations} (right plot) for an illustration. 

Both the TRec and TLin activation functions allow hidden units to use a linear rather 
than affine encoding. We shall refer to autoencoders with these activation functions 
as zero-bias autoencoder (ZAE) in the following.

\subsection{Parseval autoencoders}
For overcomplete representations orthonormality can no longer hold. 
However, if the weight vectors span the data space, 
they form a \emph{frame} (eg. \cite{kovacevic2008introduction}), 
so analysis weights ${\bm{\tilde{w}}}_i$ exist, 
such that an exact reconstruction can be written as 
\begin{equation}
\bm{r}(\bm{x}) = \sum_{k\in S(\bm{x})} \big( {\bm{\tilde{w}}}_k^\mathrm{T}\bm{x} \big) \bm{w}_k 
\label{eq:framereconstruction}
\end{equation}
The vectors ${\bm{\tilde{w}}}_i$ and $\bm{w}_i$ are in general not identical, 
but they are related through a matrix multiplication: 
$
\bm{w}_{k} = S {\bm{\tilde{w}}}_{k}. 
$
The matrix $S$ is known as frame operator for the frame $\{\bm{w}_k\}_k$ given by the 
weight vectors $\bm{w}_k$, and the set $\{\bm{\tilde{w}}_k\}_k$ is the dual frame
associated with $S$ (\cite{kovacevic2008introduction}). 
The frame operator may be the identity in which case $\bm{w}_k=\bm{\tilde{w}}_k$
(which is the case in an autoencoder with tied weights.)

Minimizing reconstruction error will make the 
frames $\{\bm{w}_k\}_k$ and $\{\bm{\tilde{w}}_k\}_k$ 
approximately duals of one another, so that Eq.~\ref{eq:framereconstruction} will 
approximately hold.
More interestingly, for an autoencoder with tied weights ($\bm{w}_k=\bm{\tilde{w}}_k$), 
minimizing reconstruction error would let the frame approximate a 
Parseval frame (\cite{kovacevic2008introduction}), such that Parseval's identity holds 
$
\sum_{k \in S(\bm{x})} \big(\bm{w}_k^\mathrm{T}\bm{x}\big)^2 = \|\bm{x}\|^2
$.


\section{Experiments}
\label{section:experiments}

\subsection{CIFAR-10}
We chose the CIFAR-10 dataset (\cite{krizhevsky2009learning}) 
to study the ability of various models to learn from high dimensional input data.  
It contains color images of size $32\times32$ pixels that are assigned 
to $10$ different classes.
The number of samples for training is $50,000$ and for testing is $10,000$. 
We consider the permutation invariant recognition task where the method is 
unaware of the 2D spatial structure of the input. We evaluated several other models 
along with ours, namely contractive autoencoder, 
standout autoencoder (\cite{ba2013adaptive}) and K-means. 
The evaluation is based on classification performance.  

The input data of size $3\times32\times32$ is contrast normalized and dimensionally reduced 
using PCA whitening retaining $99\%$ variance. We also evaluated a second method of
dimensionality reduction using PCA without whitening (denoted NW below). By whitening we mean
normalizing the variances, i.e., dividing each dimension by the square-root of the eigenvalues after PCA projection.
The number of features for each of the model 
is set to {$200,500,1000,1500,2000,2500,3000 ,3500,4000$}. 
All models are trained with stochastic gradient descent. 
For all the experiments in this section we chose a learning rate of $0.0001$ for 
a few (e.g. $3$) initial training epochs, and then increased it to $0.001$. 
This is to ensure that scaling issues in the initializing are dealt with at the outset, 
and to help avoid any blow-ups during training. 
Each model is trained for $1000$ epochs in total with a fixed momentum of $0.9$.
For inference, we use rectified linear units \emph{without bias} for all the models. 
We classify the resulting representation using logistic regression with weight decay 
for classification, with weight cost parameter estimated using cross-validation on 
a subset of the training samples of size $10000$. 

\begin{figure*}
\begin{center}
  \subfigure[TLin AE]{\includegraphics[width=0.2\linewidth]{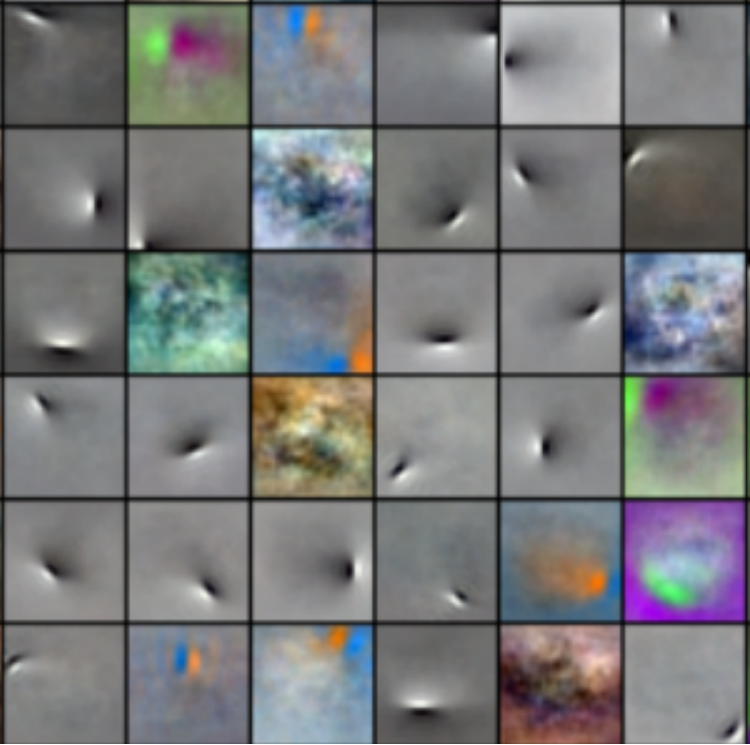}} \hspace{3mm}
  \subfigure[TRec AE]{\includegraphics[width=0.2\linewidth]{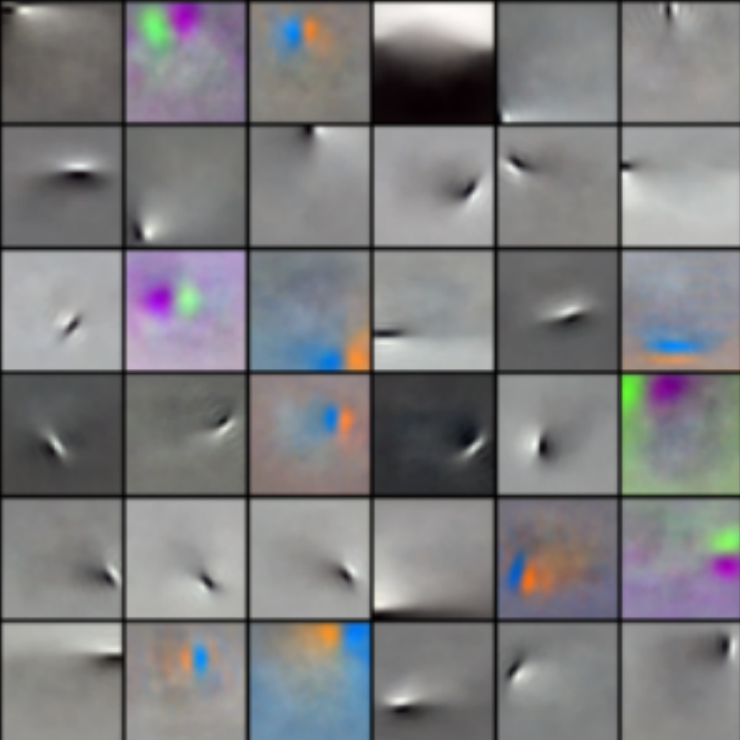}} \hspace{3mm}
  \subfigure[cAE RS=3]{\includegraphics[width=0.2\linewidth]{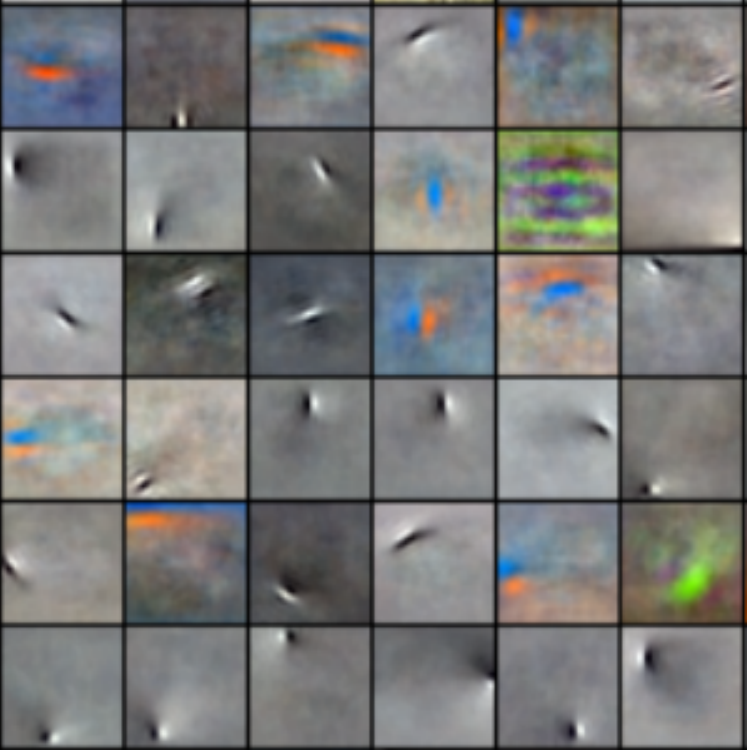}} \hspace{3mm}
  \subfigure[cAE RS=-3]{\includegraphics[width=0.2\linewidth]{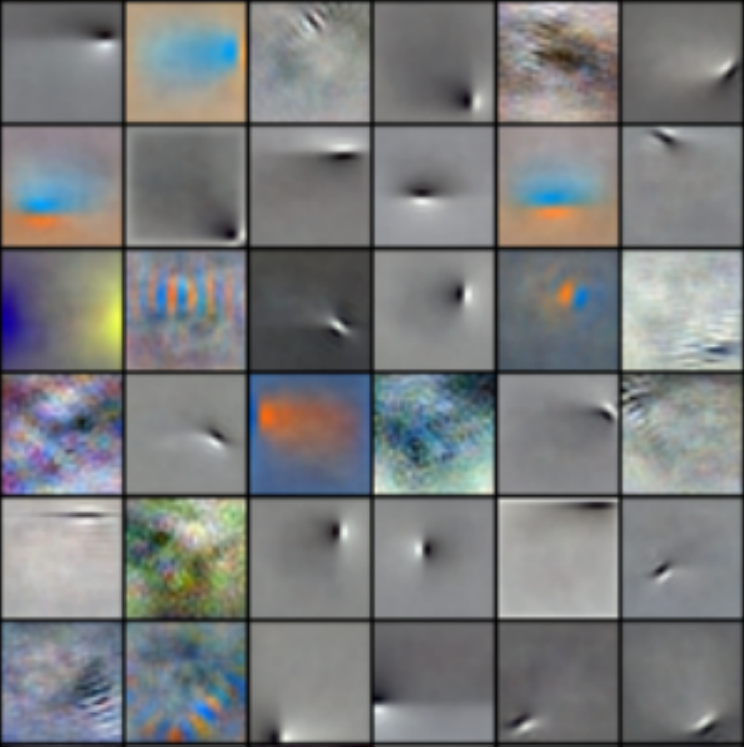}} \\

  \subfigure[TLin AE NW]{\includegraphics[width=0.2\linewidth]{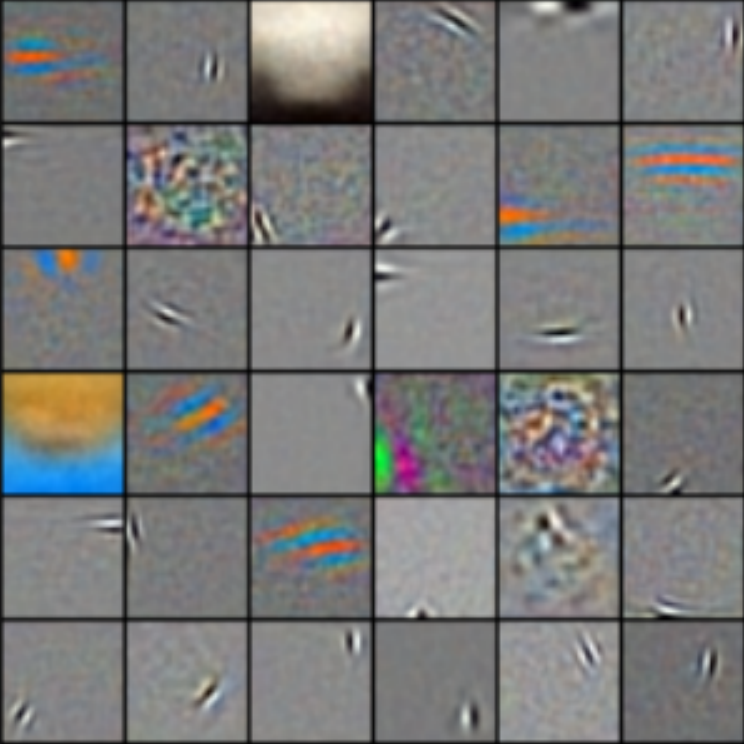}} \hspace{3mm}
  \subfigure[TRec AE NW]{\includegraphics[width=0.2\linewidth]{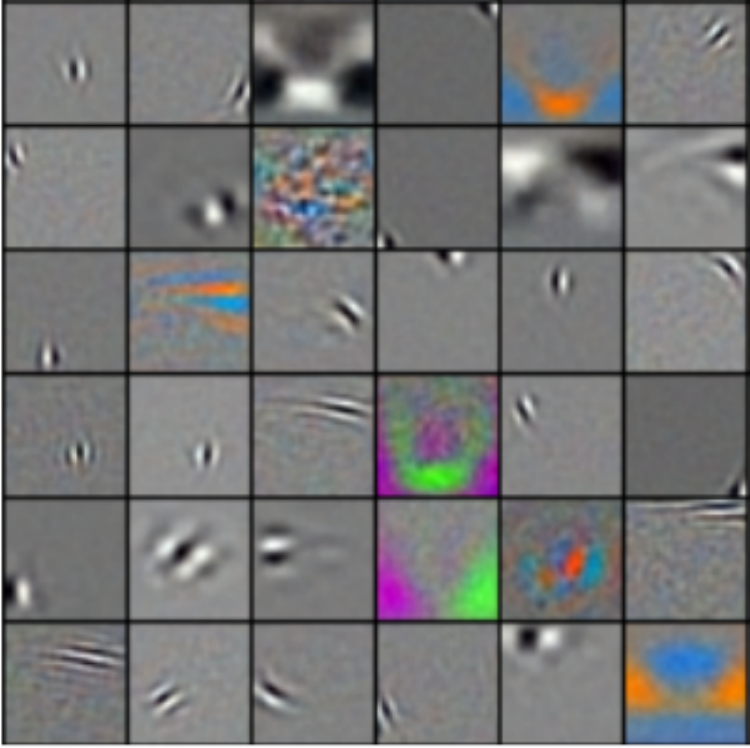}} \hspace{3mm}
  \subfigure[cAE RS=3 NW]{\includegraphics[width=0.2\linewidth]{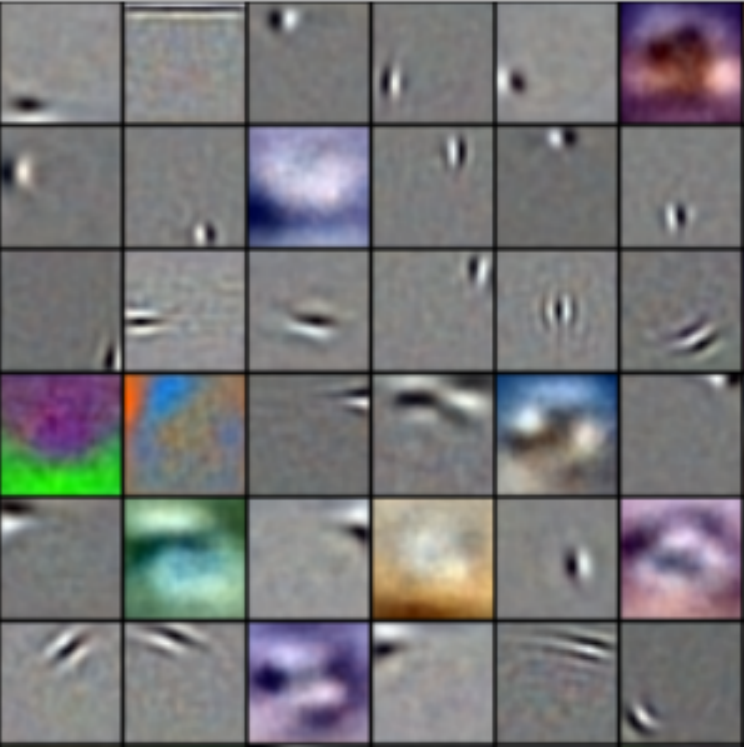}} \hspace{3mm}
  \subfigure[cAE RS=-3 NW]{\includegraphics[width=0.2\linewidth]{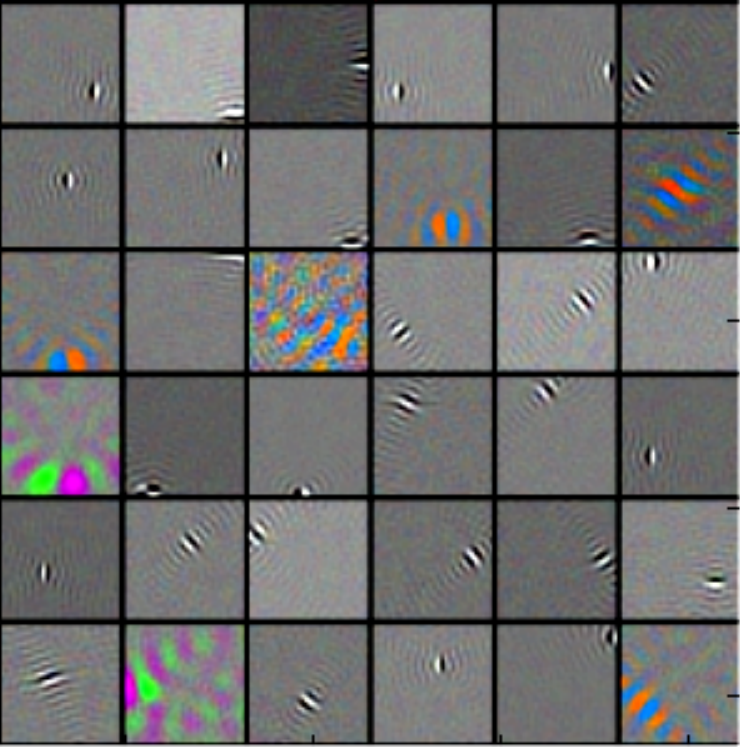}} \\
  \caption{Features of different models trained on CIFAR-10 data. Top: PCA with whitening as preprocessing. Bottom: PCA with no whitening as preprocessing. RS denotes regularization strength.}
  \label{allfilters}
\end{center}
\end{figure*}

\begin{figure}[ht]
\vskip 0.2in
\begin{center}
\begin{tabular}{cc}
\includegraphics[width=0.45\columnwidth]{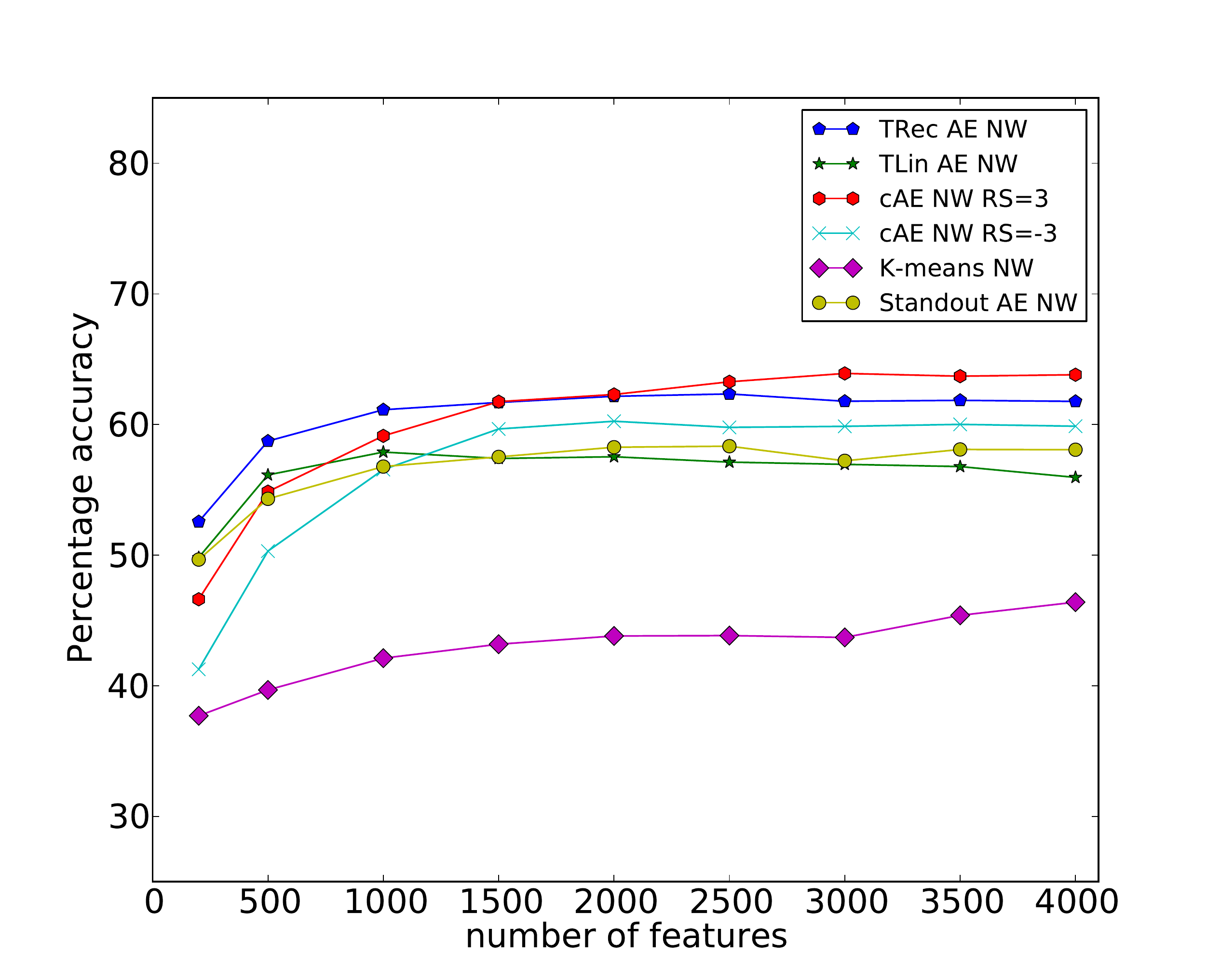}&
\includegraphics[width=0.45\columnwidth]{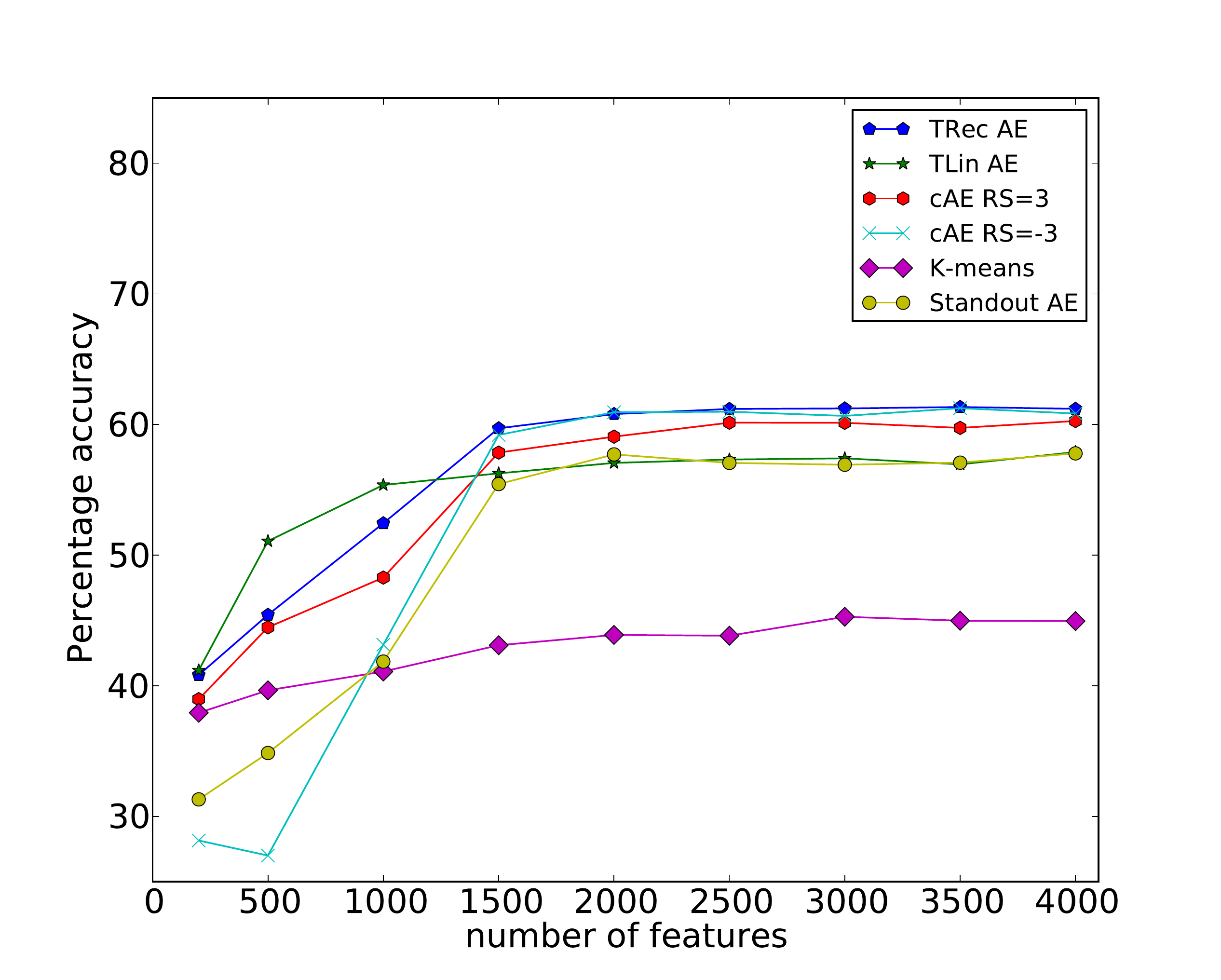}\\
\includegraphics[width=0.45\columnwidth]{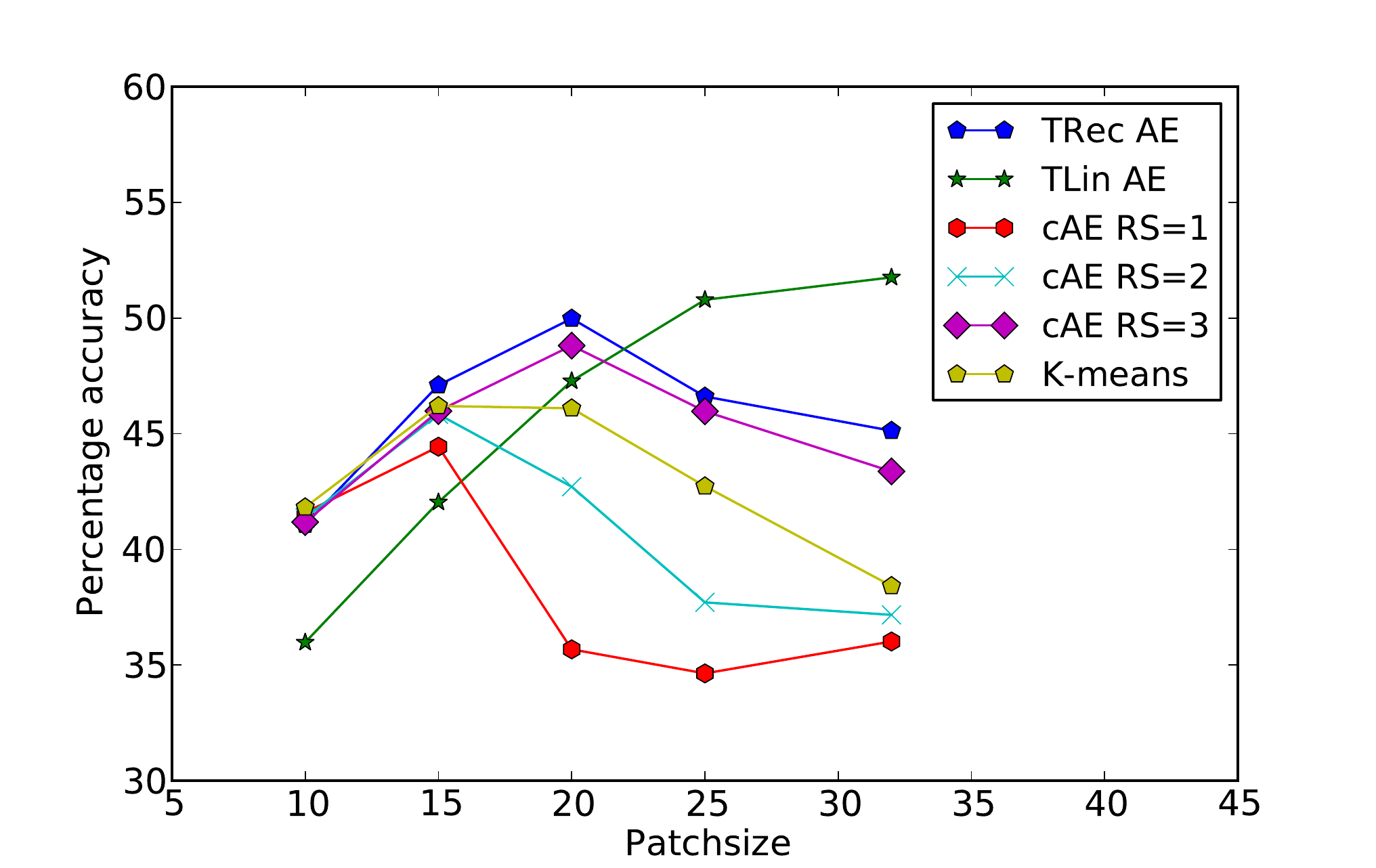} &
\includegraphics[width=0.45\columnwidth]{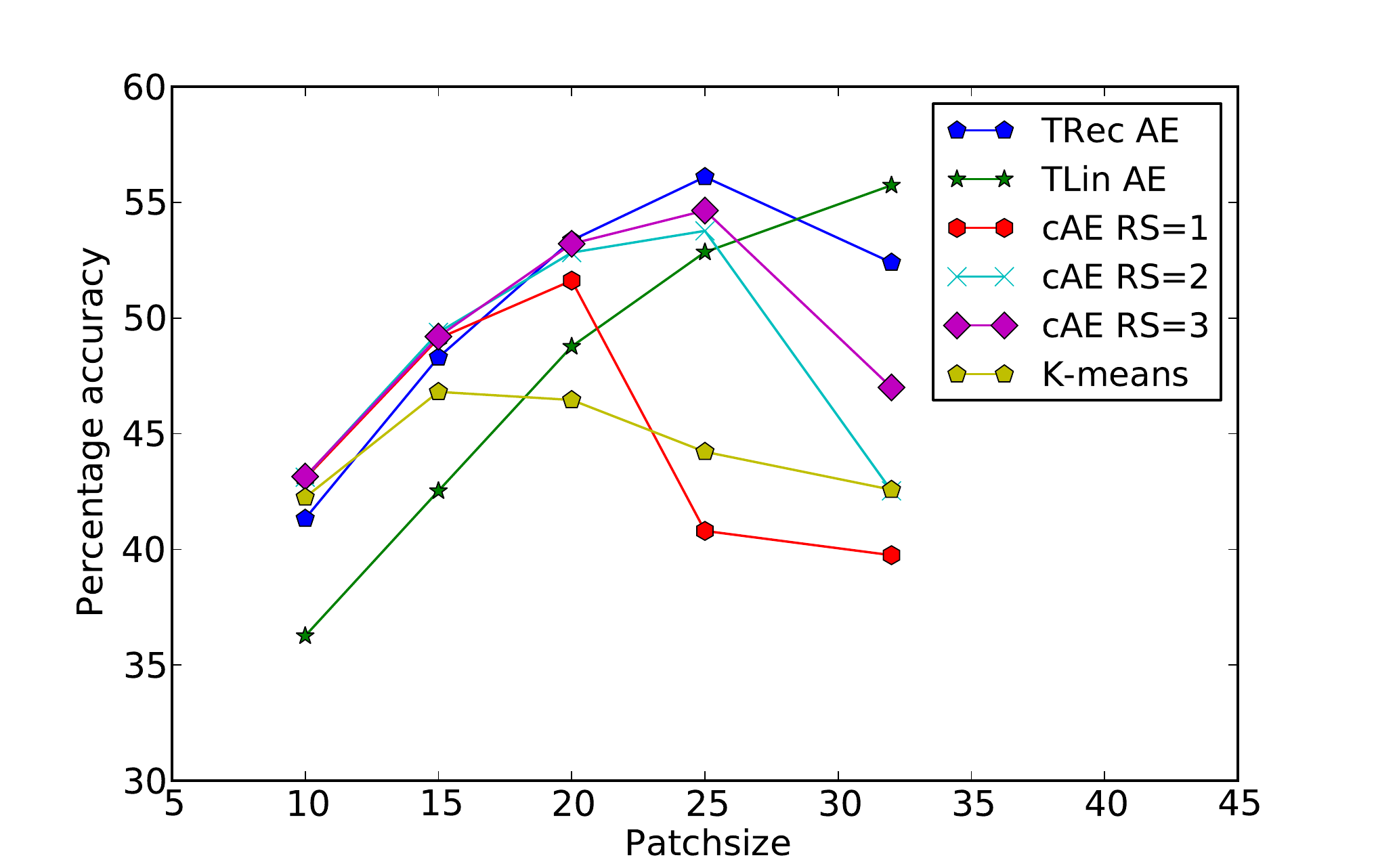}\\
\end{tabular}
\caption{Top row: Classification accuracy on permutation invariant CIFAR-10 data as a function of number of features. PCA with whitening (left) and without whitening (right) is used for preproceesing.
Bottom row: 
Classification accuracy on CIFAR-10 data for $500$ features (left) $1000$ features (right) as a function of input patch size. PCA with whitening is used for preprocessing.
}
\label{perfplot_NWWW}
\end{center}
\vskip -0.2in
\end{figure}

The threshold  parameter $\theta$ is fixed to $1.0$ for both the TRec and TLin 
autoencoder. 
For the cAE we tried the regularization strengths $1.0,2.0,3.0,-3.0$; the latter 
being ``uncontraction''.
In the case of the Standout AE we set $\alpha=1$,$\beta=-1$. 
The results are reported in the plots of Figure \ref{perfplot_NWWW}. 
Learned filters are shown in Figure~\ref{allfilters}.

From the plots in Figure \ref{perfplot_NWWW} it is observable that the results are in line 
with our discussions in the earlier sections. 
Note, in particular that the TRec and TLin autoencoders perform well even with 
very few hidden units. 
As the number of hidden units increases, the performance of the models 
which tend to ``tile'' the input space tends to improve.

In a second experiment we evaluate the impact of different input sizes on a fixed 
number of features.  
For this experiment the training data is given by image patches of 
size $P$ cropped from the center of each training image from the CIFAR-10 dataset. 
This yields for each patch size $P$ a training set of $50000$ samples 
and a test set of $10000$ samples. 
The different patch sizes that we evaluated are $10, 15, 20, 25$ as well as the original 
image size of $32$. The number of features is set to $500$ and $1000$. 
The same preprocessing (whitening/no whitening) and classification procedure as above 
are used to report performance. The results are shown in Figure~\ref{perfplot_NWWW}.


When using preprocessed input data directly for classification, the performance 
increased with increasing patch size $P$, as 
one would expect. 
Figure~\ref{perfplot_NWWW} shows that 
for smaller patch sizes, all the models perform equally well. 
The performance of the TLin based model improves monotonically as the patch size is increased. 
All other model's performances suffer when the patch size gets too large. 
Among these, the ZAE model using TRec activation suffers the least, as expected. 

We also experimented by initializing a neural network with features from the trained models. 
We use a single hidden layer MLP with ReLU units where the input to hidden weights are 
initialized with features from the trained models and the hidden to output weights
from the logistic regression models (following \cite{krizhevsky2009learning}). 
A hyperparameter search yielding $0.7$ as the optimal threshold, along with 
supervised fine tuning helps increase the best performance in the case of the TRec 
AE to $63.8$. 
The same was not observed in the case of the cAE where the performance went slightly down. 
Thus using the TRec AE followed by supervised fine-tuning with dropout regularization yields $64.1\%$ accuracy and 
the cAE with regularization strength of $3.0$ yields $63.9\%$. 
To the best of our knowledge both results beat the current state-of-the-art performance on 
the permutation invariant CIFAR-10 recognition task (cf., for example, \cite{fastfood}), 
with the TRec slightly outperforming the cAE.
In both cases PCA without whitening was used as preprocessing.
In contrast to \cite{krizhevsky2009learning} we do not train on any extra data, so 
none of these models is provided with any knowledge of the task beyond the preprocessed 
training set.

\subsection{Video data}
An dataset with very high intrinsic dimensionality are videos that show transforming 
random dots, as used in \cite{factoredGBM} and subsequent work: 
each data example is a vectorized video, whose first frame is a random image and whose 
subsequent frames show transformations of the first frame. 
Each video is represented by concatenating the vectorized frames into a large vector. 
This data has an intrinsic dimensionality which is at least as high as the dimensionality 
of the first frame. So it is very high if the first frame is a random image. 

It is widely assumed that only bi-linear models, such as \cite{factoredGBM} and related models, 
should be able to learn useful representations of this data. 
The interpretation of this data in terms of high intrinsic dimensionality suggests that a
simple autoencoder may be able to learn reasonable features, as long as it uses a 
linear activation function so hidden units can span larger regions. 

\begin{figure}[h!]
\begin{center}
\begin{tabular}{cc}
\hspace{-7pt}
  \includegraphics[width=0.40\linewidth]{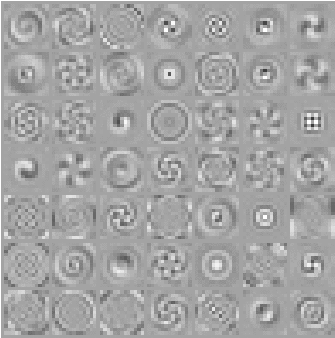} &
  \includegraphics[width=0.40\linewidth]{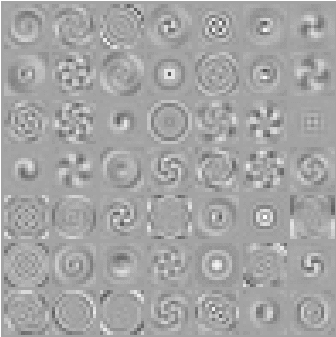} 
\end{tabular}
\begin{tabular}{lc}
\hline
  Model			&Average precision \\
  \hline
  TRec AE 		&50.4 \\
  TLin AE		&49.8 \\
  covAE	\cite{gated}	&43.3 \\
  GRBM \cite{convGBM}	&46.6 \\
  K-means		&41.0 \\
  contractive AE	&45.2\\
\hline
\end{tabular}
\end{center}
\caption{Top: Subset of filters learned from rotating random dot movies (frame 2 on the left, frame 4 on the right).
Bottom: Average precision on Hollywood2.}
\label{figure:rotationfilters}
\end{figure}

We found that this is indeed the case by training the ZAE on rotating 
random dots as proposed in \cite{factoredGBM}. The ZAE model with $100$ hiddens is trained 
on vectorized $10$-frame random dot videos with $13\times13$ being the
size of each frame. Figure~\ref{figure:rotationfilters} depicts filters learned and
shows that the model learns to represent the structure in this data by developing phase-shifted 
rotational Fourier components as discussed in the context of bi-linear models. 
We were not able to learn features that were distinguishable from noise with the cAE, which 
is in line with existing results (eg. \cite{factoredGBM}).

We then chose activity recognition to perform a quantitative evaluation of this observation. 
The intrinsic dimensionality of real world movies is probably lower than that of random 
dot movies, but higher than that of still images. 
We used the recognition pipeline proposed in \cite{QuocISA,KondaMM13} and 
evaluated it on the Hollywood2 dataset \cite{marszalek09}.
The dataset consists of $823$ training videos and $884$ test videos with $12$ classes 
of human actions.
The models were trained on PCA-whitened input patches of size $10\times16\times16$ cropped randomly
from training videos. 
The number of training patches is $500,000$. 
The number of features is set to $600$ for all models.

In the recognition pipeline, sub blocks of the same size as the patch size are cropped 
from $14\times20\times20$ “super-blocks”, using
a stride of $4$. Each super block results in $8$ sub blocks. 
The concatenation of sub block filter responses is dimensionally reduced by 
performing PCA to get a super block descriptor, 
on which a second layer of K-means learns a vocabulary of spatio-temporal words, 
that get classified with an SVM (for details, see \cite{QuocISA,KondaMM13}). 

In our experiments we plug the features learned with the different models 
into this pipeline.  
The performances of the models are reported in Figure \ref{figure:rotationfilters} (right). 
They show that the TRec and TLin autoencoders clearly outperform the more 
localized models. Surprisingly, they also outperform more sophisticated gating models, 
such as \cite{factoredGBM}. 

\subsection{Rectified linear inference}
\label{sec:RectlinInf}
In previous sections we discussed the importance of (unbiased) rectified linear inference. Here we experimentally show that using 
rectified linear inference yields the best 
performance among different inference schemes. 
We use a cAE model with a fixed number of hiddens trained on CIFAR-10 images, and 
evaluate the performance of 
\begin{enumerate} \itemsep1pt \parskip0pt \parsep0pt
\item Rectified linear inference with bias (the natural preactivation for the unit): $[W^TX+b]_+$ 
\item Rectified linear inference without bias: $[W^TX]_+$ 
\item natural inference: $\mathrm{sigmoid}(W^TX+b)$ 
\end{enumerate}
The performances are shown in Figure \ref{figure:negbiases} (right), 
confirming and extending the results presented in \cite{coatessinglelayer,ICML2011Saxe_551}.


\section{Discussion}
Quantizing the input space with tiles proportional in quantity to the 
data density is arguably the best way to represent data given enough training data and 
enough tiles, because it allows us to approximate any function reasonably well 
using only a subsequent linear layer. 
However, for data with high intrinsic dimensionality and a limited number of hidden 
units, we have no other choice than to summarize regions using responses that are invariant 
to some changes in the input. 
Invariance, from this perspective, is a necessary evil and not a goal in itself. But it is increasingly important for increasingly high dimensional inputs. 

We showed that linear not affine hidden responses allow us to get invariance,  
because the density defined by a linear autoencoder is a superposition 
of (possibly very large) regions or subspaces. 


After a selection is made as to which hidden units are active 
for a given data example, linear coefficients are used in the reconstruction. 
This is very similar to the way in which gating and square pooling models
(eg., \cite{OlshausenBilinear,Memisevic07,factoredGBM,mcrbm,QuocISA,courville2011spike}) define 
their reconstruction: 
The response of a hidden unit in these models is defined by multiplying the filter 
response or squaring it, followed by a non-linearity. 
To reconstruct the input, the output of the hidden unit 
is then multiplied by the filter response itself, making the model bi-linear. 
As a result, reconstructions are defined as the sum of feature vectors, weighted by 
\emph{linear} coefficients of the active hiddens. 
This may suggest interpreting the fact that these models work well on videos 
and other high-dimensional data as a result of using linear, zero-bias hidden units, too. 

\section*{Acknowledgments} This work was supported by an NSERC Discovery grant, a Google faculty research award, and the German Federal Ministry of Education and Research (BMBF) in the project 01GQ0841 (BFNT Frankfurt).

{\small
\bibliography{zae}
\bibliographystyle{iclr2015}
}

\end{document}